\RequirePackage{amsmath}

\documentclass{svjour3} 

\usepackage[version=3]{mhchem} 
\usepackage[T1]{fontenc}       
\usepackage{verbatim}
\usepackage{graphicx}
\usepackage{url}
\usepackage{subfigure}
\usepackage[sort, numbers]{natbib}
\usepackage{fix-cm}

\begin{document}
\title{High-Dimensional Materials and Process Optimization using Data-driven Experimental Design with Well-Calibrated Uncertainty Estimates}
\titlerunning{Data-driven Experimental Design} 
\author{Julia Ling* \and Maxwell Hutchinson* \and Erin Antono \and Sean Paradiso \and Bryce Meredig}
\institute{
*These authors contributed equally to this work\\ 
Citrine Informatics, Redwood City, California, jling@citrine.io\\
}
\date{Received: 4/12/17}
\authorrunning{Ling et al.}

\maketitle
\begin{abstract}
The optimization of composition and processing to obtain materials that exhibit desirable characteristics has historically relied on a combination of domain knowledge, trial and error, and luck. 
We propose a methodology that can accelerate this process by fitting data-driven models to experimental data as it is collected to suggest which experiment should be performed next. 
This methodology can guide the practitioner to test the most promising candidates earlier, and can supplement scientific and engineering intuition with data-driven insights. 
A key strength of the proposed framework is that it scales to high-dimensional parameter spaces, as are typical in materials discovery applications. 
Importantly, the data-driven models incorporate uncertainty analysis, so that new experiments are proposed based on a combination of \textit{exploring} high-uncertainty candidates and \textit{exploiting} high-performing regions of parameter space. 
Over four materials science test cases, our methodology led to the optimal candidate being found with three times fewer required measurements than random guessing on average.

\end{abstract}

\section{Introduction}

Because of the time intensity of performing experiments and the high dimensionality of many experimental design spaces, exploring an entire parameter space is often prohibitively costly and time-consuming.
The field of \textit{sequential learning} (SL) is concerned with choosing the parameter settings for an experiment or series of experiments in order to either maximize information gain or move toward some optimal parameter space.
In general, these parameter settings encompass everything from measurement procedures to physical test conditions and test specimen characteristics.
Sometimes also called \textit{optimal experimental design} or \textit{active learning}, sequential learning can be used by the experimenter to decide which experiment to perform next to most efficiently explore the parameter space.

Traditional design-of-experiment approaches are typically applied to relatively low-dimensional optimization problems. 
For example, the Taguchi method relies on performing a set of experiments to create a complete basis of orthogonal arrays, which requires gridding the input parameters \textit{a priori} into a set of plausible values~\cite{Roy2010}.  
Fisher's \textit{analysis of variance} approach, which decomposes the variance of a response variable into the contributions due to different input parameters, also assumes that the input parameters are discrete~\cite{Fisher1921}.
These approaches can be powerful in certain applications, such as reducing process variability, but do not scale to larger-dimensional, more exploratory scenarios with real-valued, inter-dependent input parameters.
Many SL approaches rely on Bayesian statistics to inform their choice of experiments~\cite{Chernoff1959, Chaloner1995, Cohn1996}.
In this case, the experimental response function (\textit{i.e.} the quantity being measured in the experiment) $f(\vec{x})$ is estimated by a surrogate model $\hat{f}(\vec{x} ; \vec{\theta})$, where $\vec{x}$ are the experimental parameters and $\vec{\theta}$ are the surrogate model parameters.
In the Bayesian setting, the experimental data are used to estimate an a posteriori joint probability distribution function for the model parameters.
The Bayesian approach has two main strengths.
First, it provides uncertainty bounds on the estimated model response $\hat{f}(\vec{x} ; \vec{\theta})$.
These uncertainty bounds can be used to inform the choice of experiments.
The second advantage of Bayesian optimization, as opposed to gradient-based optimization, is that it uses all previous measurements to inform the next step in the optimization process, resulting in an efficient use of the collected data~\cite{Martinez2014}.
On the other hand, Bayesian methods often struggle in high-dimensional spaces due to the curse of dimensionality in constructing a joint probability distribution function between many parameters.
High dimensional spaces are typically handled by first applying dimension reduction techniques~\cite{Shan2010}.

Recently, there has been increasing interest in SL approaches for applications in materials science.
\citet{Wang2015} applied SL to the design of nanostructures for photoactive devices.
They proposed a Bayesian SL method that suggested experiments in batches.
They used Monte Carlo sampling to estimate the dependence of the response function on their two experimental parameters.
Their approach was shown to optimize the nanostructure properties with fewer trials than a greedy approach that did not leverage uncertainty information.
\citet{Aggarwal2015}  applied Bayesian methods to two different applications in materials science: characterizing a substrate under a thin film, and selecting between models for the trapping of Helium impurities in metal composites.
~\citet{Ueno2016} presented a Bayesian optimization framework which they applied to the case of determining the atomic structure of crystalline interfaces.
\citet{Xue2017} investigated the use of SL for discovering shape memory alloys with high transformation temperatures.  
They used a simple polynomial regression on three material parameters to drive their predictions.
\citet{Dehghannasiri2017} also proposed the use of SL for the discovery of shape memory alloys.
They used the mean objective cost of uncertainty algorithm, an SL approach that performs robust optimization to incorporate the cost of uncertainty.
In the test case of designing shape memory alloys with low energy dissipation, their approach was shown to require fewer trials than either random guessing or greedy optimization.
These studies highlighted the significant promise of SL for reducing the number of experiments required to achieve specified performance goals in materials science applications.
However, these SL approaches were all evaluated on case studies with five or fewer degrees of freedom.

In materials science, it is not always straightforward to describe an experimental design in terms of a small number of real-valued parameters.
For example, in trying to determine a new alloy with specific characteristics, how should the alloy formula be parametrized? 
In an effort to discover new Heusler compounds, \citet{Oliynyk2016} parametrized the chemical formula in terms of over fifty different real-valued and categorical features that could be calculated directly from the formula.
Such high-dimensional parameter spaces demand a different SL strategy.
In this paper, we present an SL approach that uses random Forests with Uncertainty Estimates for Learning Sequentially (FUELS) to scale to high-dimensional (> 50 dimensions) applications.
It is worth noting that the focus here is on high-dimensional input parameter spaces, not on multi-objective optimization, which is beyond the scope of the current study.
We evaluate the performance of the FUELS framework on four different test cases from materials science applications: one in magnetocalorics, one in superconductors, another in thermoelectrics, and the fourth in steel fatigue strength.
There are two main innovations presented in this paper.  
The first is the implementation of robust uncertainty estimates for random forests, validated for the four test cases.  
These uncertainty estimates are critical not only for the application of SL, but also for making data-driven predictions in general.  
For example, if a model for steel fatigue strength predicts only a raw number without uncertainty, it is impossible to know if the model is confident in that prediction or is extrapolating wildly.  
In this case, it would not be clear whether more data needed to be collected before the model could be trusted as part of the design process.
Because of the many sources of uncertainty in materials science and the reliability-driven nature of design specifications, it is key that as data-driven models rise in popularity, methods for quantifying their uncertainty are developed and evaluated.
The second major innovation is the application of these random forests with uncertainty bounds to high-dimensional SL test cases in materials science.
We demonstrate their utility as a practical experimental design tool for materials and process optimization that significantly reduces the number of experiments required.

\section{Methodology}\label{MethodologySec}

\begin{figure}[t]
\begin{center}
\includegraphics [width=100mm]{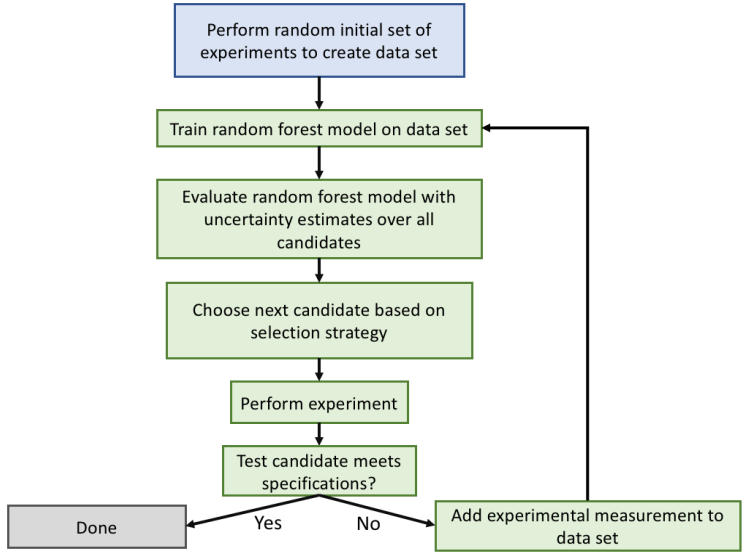}
\end{center}
\caption{Schematic of the proposed FUELS framework} 
\label{OED_Schematic} 
\end{figure}

The proposed FUELS methodology is built on a random forest model that includes model uncertainty estimation at each candidate test point.
The algorithm suggests the next candidate to test based on maximizing some selection function over the unmeasured candidates.
Figure~\ref{OED_Schematic} shows a schematic of the FUELS framework.
The following subsections describe the random forest algorithm, the uncertainty analysis procedure and evaluation, and the candidate selection strategies.

\subsection{Random forests with uncertainty}

Random forests are composed of an ensemble of decision trees \cite{Breiman2001, ho1998}.
Decision trees are simple models that recursively partition the input space and define a piece-wise function, typically constant, on the partitions.
Single decision trees often have poor predictive power for non-linear relations or noisy data sets~\cite{Breiman2001}.
In random forests, these weaknesses are overcome by using an ensemble of decision trees, each one fit to a different random draw, with replacement, of the training data set.
For a given test point, the predictions of all the trees in the forest are aggregated, usually by taking the mean.

The uncertainty estimates used in FUELS build on work by ~\citet{efron2012model} and \citet{Wager2014}, differing principally in the inclusion of an explicit bias term.
The variability of the tree predictions and the correlation of tree predictions with the inclusion of points in their randomly drawn training sets are used to estimate the random forest uncertainty.
The uncertainty is estimated as sum of contributions from each training point:
\begin{equation}
	\sigma(\vec{x}) = \sqrt{\left(\sum\limits_{i=1}^S \text{max}\left[\sigma_i^2(\vec{x}), \omega\right]\right) + \tilde\sigma^2(\vec{x})},
	\label{uq_eqn}
\end{equation}
where 
$\sigma_i^2(\vec{x})$ is the sample-wise variance at test point \vec{x} due to training point $i$,
$\omega$ is the noise threshold in the sample-wise variance estimates, and
$\tilde\sigma(\vec{x})$ is an explicit bias function, to be discussed later.
In this work, the noise threshold is set to $\omega = \left|\min_i \sigma^2(\vec{x}_i) \right|$, the magnitude of the minimum variance over the training data.

The sample-wise variance is defined as the average of the jackknife-after-bootstrap and infinitesimal jackknife variance estimates with a Monte Carlo sampling correction~\cite{Wager2014}:
\begin{equation}
	\sigma^2_i(\vec{x}) = \text{Cov}_j\left[n_{i,j}, t_j(\vec{x})\right]^2 + \left[\overline{t}_{-i}(\vec{x}) - \overline{t}(\vec{x})\right]^2 - \frac{e v}{T},
	\label{ij_eqn}
\end{equation}
where $\text{Cov}_j$ is the covariance computed over the tree index $j$,
$n_{i,j}$ is the number of instances of the training point $i$ used to fit tree $j$,
$t_j(\vec{x})$ is the prediction of the $j$th tree,
$\overline{t}_{-i}(\vec{x})$ is the average over the trees that were not fit on sample $i$,
$\overline{t}(\vec{x})$ is the average over all trees,
$e$ is Euler's number,
$v$ is the variance over all the trees, and
$T$ is the number of trees.

The sample-wise variance is effective at capturing the uncertainty due to the finite size of the training data.
It can, however, underestimate uncertainty due to noise in the training data or unmeasured degrees of freedom.
For these reasons, we amended the sample-wise variance with the explicit bias model $\tilde\sigma(\vec{x})$, which should be chosen to be very simple to avoid over-fitting.
Here, $\tilde\sigma(\vec{x})$ is a single decision tree limited to depth $\log_2(S)/2$, where $S$ is the number of training points.
The random forests and uncertainty estimates used in this study are available in the open source Lolo scala library~\cite{Lolo}.

\subsection{Evaluation of Uncertainty Estimates}

\begin{figure}[t]
\begin{center}
\subfigure[Normalized by FUELS uncertainty estimate]{\label{} \includegraphics[width=130mm]{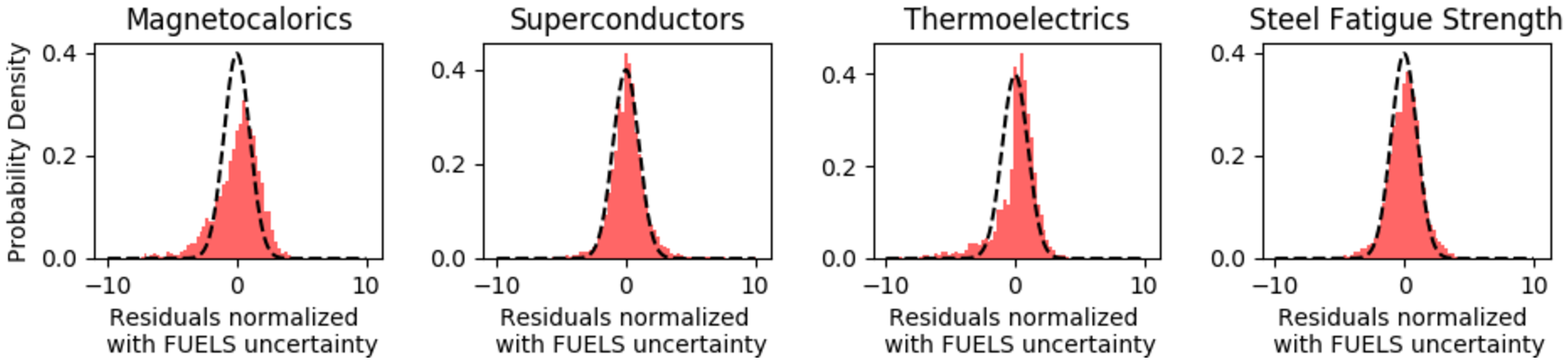}}\\
\subfigure[Normalized by out-of-bag error]{\label{} \includegraphics[width=130mm]{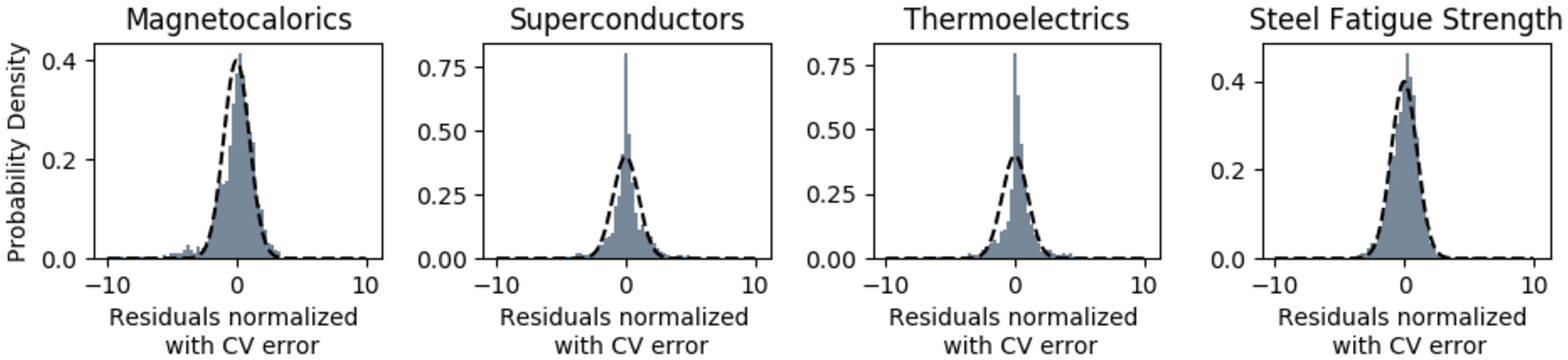}}
\caption{Probability densities of normalized residuals computed via 8-fold cross validation for each of the 4 cases. 
The residuals are normalized by (a) the FUELS uncertainty estimate $\sigma (\vec{x})$ (see Eq.~\ref{ij_eqn}) and (b) the root mean square out-of-bag error, which is equivalent to removing the Jackknife-based uncertainty and using a constant model for the explicit bias.
The unit normal distribution, representing perfectly calibrated uncertainty estimates, is shown for reference with the dashed black line.}
\label{residuals_hist_fig} 
\end{center}
\end{figure}

We evaluated these uncertainty estimates on the four data sets that will be explored as test cases in this paper.
These data sets include a magnetocalorics data set~\cite{Bocarsly2017}, a superconductor data set compiled by the internal Citrine team, a thermoelectrics data set~\cite{Sparks2016}, and a steel fatigue strength data set~\cite{Agrawal2014}, which will all be described in more detail in Section~\ref{ResultsSec}.
Models were trained to predict the magnetic deformation, superconducting critical temperature, figure of merit ZT, and fatigue strength, respectively, on these four data sets.
The models were evaluated via eight-fold cross-validation over 16 trials and the validation error was compared to the combined uncertainty estimates.

Figure~\ref{residuals_hist_fig} shows the probability densities of the normalized residuals for the four test cases.
The normalized residuals $r_{n}$ are given by $r_{n} = \frac{\hat{f}(\vec{x}) - f(\vec{x})} {\sigma (\vec{x})}$.
In other words, $r_{n}$ is the difference between the predicted and actual value, divided by the uncertainty estimate.
If the uncertainty estimates were perfectly well calibrated and the samples in the data set were independently distributed, then the normalized residuals would follow a Gaussian distribution with zero mean and unit standard deviation. 
As the histograms show, the distributions of the normalized residuals are roughly normal, albeit with heavier tails than a normal distribution.
Figure~\ref{residuals_hist_fig} also shows the residuals normalized by the root mean square out-of-bag error, which is equivalent to removing the jackknife-based contributions to the uncertainty and using the simplest explicit bias model, i.e. a constant function.
In this context, the out-of-bag error on a training example refers to the average error of predictions made using the subset of decision trees that were not trained on that particular training example.
The root mean squared out-of-bag error is analogous to the conventional cross-validation error, which provides a constant error estimate for all test points.
The figure demonstrates that the root mean square out-of-bag error is not a well-calibrated uncertainty metric; it drastically over-estimates the error for a large fraction of the points in the thermoelectrics and superconductor test cases, as demonstrated by the large difference between the standard normal distribution and the residuals near 0 in Fig.~\ref{residuals_hist_fig}.

The heavy tails shown in Fig.~\ref{residuals_hist_fig}(a) are not unexpected, since the current estimates cannot fully account for all sources of uncertainty, such as uncertainty due to contributions that cannot be explained with the given feature set, i.e. ``unknown unknowns.''
For example, if the target function is conductivity and different data points were acquired at different temperatures, but those temperatures were not measured and added to the training data, then the missing information can cause the uncertainty estimates to be unreliable.
Such unknown unknowns are likely responsible for the few outliers seen in Fig.~\ref{residuals_hist_fig}.
Nevertheless, this examination of the uncertainty estimates shows that they give a reasonable representation of the random forest model uncertainty.
This uncertainty estimation procedure is of broad utility for providing quantitative uncertainty bounds for data-driven random forest models, and was used in the present study for the purpose of SL.

\subsection{FUELS Framework}

The schematic in Fig.~\ref{OED_Schematic} outlines how the random forest and uncertainty estimates are applied to SL.
In this study, it is assumed that the goal of SL is to determine the optimal material processing and composition from a list of candidate options using the fewest possible number of experiments.
Optimality is based on maximizing (or minimizing) some material property, such as the critical temperature for superconductivity.

The first step in the SL framework is to evaluate the response function for an initial set of test candidates in order to fit a random forest model for the response function.
In this study, this initial set of test candidates consisted of 10 randomly selected materials from the set of candidates.
Future work will investigate the optimal size of this initial set, as well as explore sampling strategies other than random sampling for their selection.

\begin{figure}[t]
\begin{center}
\includegraphics [width=90mm]{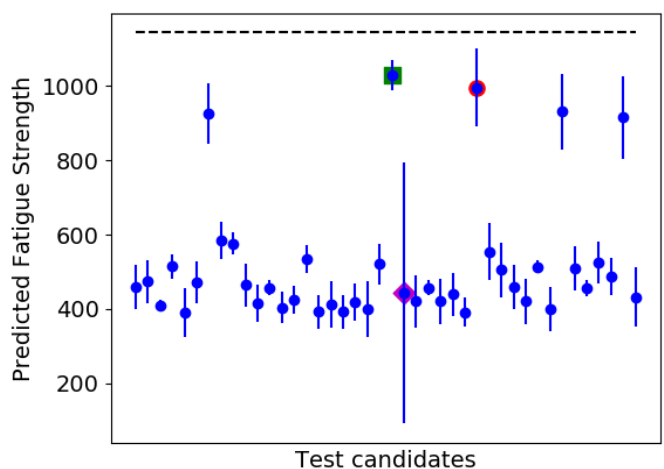}
\end{center}
\caption{Example of the MLI, MEI, and MU strategies for selecting the next point to test.  Each point represents a different candidate test point, with the value given by the random forest model prediction with uncertainty bars.  The dashed black line indicates the performance of the best candidate that has already been tested.  MLI chooses the point outlined with a red circle to test next, MEI chooses the point outlined in a green square to test next, and MU chooses the point outlined in the magenta diamond to test next.} 
\label{MLI_explanation_fig} 
\end{figure}

Once a random forest model has been fit, it is evaluated for each of the un-measured candidates.
Three different strategies for selecting the next candidate were assessed: Maximum Expected Improvement (MEI), Maximum Uncertainty (MU), and Maximum Likelihood of Improvement (MLI).
The MEI strategy simply selects the candidate with the highest (or lowest, for minimization) target value.  
The MU strategy selects the candidate with the greatest uncertainty, entirely independently of its expected value.
The MLI strategy selects the candidate that is the most likely to have a higher (or lower, for minimization) target value than the best previously measured material.  
This strategy uses the uncertainty estimates from Eq.~\ref{uq_eqn} and assumes that the uncertainty for a given prediction obeys a Gaussian distribution.
While MLI and MEI are both greedy optimization strategies, the MLI strategy typically favors evaluating candidates with high uncertainty, leading to more exploration of the search space.

Figure~\ref{MLI_explanation_fig} shows an example comparing MLI, MEI, and MU.
This example comes from one iteration of SL in the steel fatigue strength test case, in which the goal is to determine the candidate with the maximum fatigue strength.
In determining the next candidate to test, MEI would choose the candidate for which the random forest had predicted the highest value, whereas MLI selects a candidate with higher uncertainty in this case, since it has a higher probability of surpassing the best previously measured candidate.
MU selects a point with lower expected performance but higher uncertainty than those selected by MEI and MLI.
Because the uncertainty estimates tend to be higher in regions of parameter space that have not yet been explored, the MLI strategy tends to favor exploring the parameter space more fully than MEI.
MU is a purely exploration-based strategy, where the candidate with the largest uncertainty estimate is selected to be tested next.
In the context of optimization, exploitative strategies search the space near a top-performing candidate to find a local optimum.
Explorative strategies search regions farther from previously tested candidates to try to find the global optimum.
These three strategies were all applied on the four test cases to evaluate their performance.
The SL process stops when a candidate is tested which exhibits the desired performance.

\section{Results on Test Cases}\label{ResultsSec}

The FUELS framework was evaluated for four different application cases from materials science: magnetocalorics, superconductors, thermoelectrics, and steel fatigue strength.
In each of these test cases, a data set was already publicly available on Citrination with a list of potential candidate materials and their previously reported target values\footnote{https://links.citrination.com/magnetocaloric-benchmark}\footnote{https://links.citrination.com/superconductor-benchmark}\footnote{https://links.citrination.com/thermoelectric-benchmark}\footnote{https://links.citrination.com/steel-fatigue-strength-benchmark}.
The goal of the FUELS process was to identify the candidate with the maximal value of the response function, using measurements of the response from the fewest number of candidates possible.
  It should be noted that because the test sets consist of candidates that have been previously measured, there is potential sample bias in these data sets: high-performance materials are more likely to have measurements available in public data sets.
This sample bias means that there are fewer obvious bad candidates for the SL model to pass over, in effect making the problem more difficult.
Future work will test this SL methodology on a case study for which the target values are not previously available. 

In each test case, the FUELS methodology was run 30 times for each of the three strategies (MLI, MEI and MU), in order to collect statistics on the number of measurements required to find the optimal candidate.
The FUELS methodology was benchmarked against two other algorithms: random guessing and the COMBO Bayesian SL framework proposed by \citet{Ueno2016}.
In random guessing, the next candidate was selected randomly from the pool of candidates that had not been previously measured.
As a result, the number of evaluations required follows a uniform distribution over the range of the data set size.
In the COMBO strategy, a Gaussian process for the target variable is constructed and is queried to determine the next candidate to test.
Unlike the FUELS approach, COMBO uses Bayesian methods to obtain uncertainty estimates by propagating uncertainty in model parameters through to the model predictions.
COMBO uses state-of-the-art algorithms for scalability to large data sets and is a challenging benchmark strategy against which to compare the performance of FUELS.

\subsection{Magnetocalorics Test Case}
\subsubsection{Problem Description}
A magnetocaloric material exhibits a decrease in entropy when a magnetic field is applied at temperatures near its Curie temperature.
This property can be exploited for magnetization refrigeration, with larger entropy changes enabling more efficient cooling.
\citet{Bocarsly2017} showed that the entropy change of a material is strongly correlated with its magnetic deformation, a property that can be calculated via density functional theory (DFT).
They presented a reference data set of 167 candidates for magnetocaloric behavior for which the magnetic deformation had already been calculated.

In this test case, the FUELS framework was used to identify the candidate with the highest value of magnetic deformation.
If the FUELS process can efficiently identify candidates with large values of magnetic deformation, then it could be used to more efficiently determine which DFT calculations to perform.
These DFT calculations could then, in turn, be used to identify the most promising candidates for experimental testing. 

The free parameter in this test case is the material formula.
Because the material formula is not in itself a continuously-varying real-valued variable, it was parameterized in terms of 54 real-valued features that could be calculated directly from the formula~\cite{Ward2016}.
These features included quantities such as the mean electron affinity of the atoms in the compound, the orbital filling characteristics, and the mean ionization energy.
These 54 features composed the inputs to the FUELS algorithm, and the target was the magnetic deformation.

\subsubsection{Results}

\begin{figure}[t]
\begin{center}
\includegraphics [width=120mm]{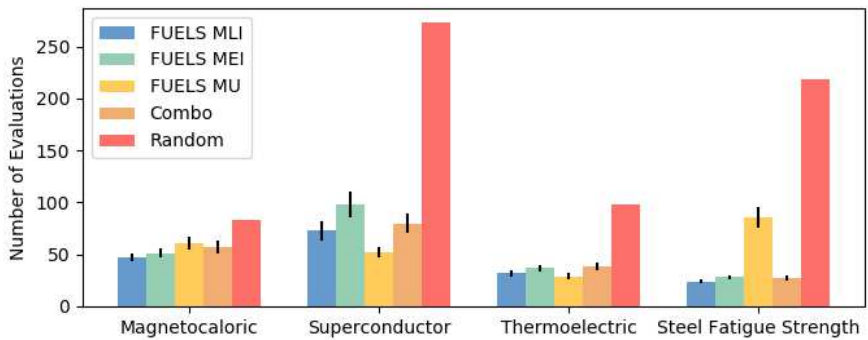}
\end{center}
\caption{
Sample mean of the number of steps required to find the optimal candidate for the four test cases using different strategies.  The magnetocaloric, superconductor, thermoelectric, and steel fatigue strength data sets had 167, 546, 195, and 437 potential candidates, respectively.
} 
\label{steps_fig} 
\end{figure}

Figure~\ref{steps_fig} shows the mean number of evaluations required to identify the candidate with the highest magnetic deformation in the test set.  
These values are also tabulated in Table~\ref{steps_table}.
The error bars represent the uncertainty in the sample mean, and were calculated as the standard deviation of the number of required measurements over the 30 trials, divided by $\sqrt{30}$.
For the random guessing strategy, the mean was given by that of a uniform distribution over the size of the data set.
As is shown in this table, FUELS (with all three strategies) and COMBO methods were all able to identify the best candidate with significantly fewer evaluations than random guessing. 
FUELS MLI finds the optimal candidate in just over half the number of evaluations as random guessing.

\begin{table}[b]
  \centering
  \caption{Sample mean and uncertainty in the sample mean at one standard deviation, for the number of steps required for different SL strategies to find the optimal candidate. }
  \label{steps_table}
  \begin{tabular}{ | l | c | c |c |c |c |c |c|}
\hline
   & Data size & \# inputs & FUELS MLI & FUELS MEI & FUELS MU& COMBO & random \\\hline
   Magnetocaloric 	& 167  & 	54& 	$47 \pm 3$ 	& $51 \pm 4 $& $61\pm 6$ 	& $57 \pm 6$ &  $84	$\\\hline
   Superconductor 	&  546 &	54& 	$73 \pm 9$ 	& $98 \pm 12 $& $52 \pm 5$	& $80 \pm 9$ & $273 $\\\hline
   Thermoelectric 	&  195 &	56&	$32 \pm 3$	& $37 \pm 3 $& $29 \pm 3$	& $38 \pm 4$ & $ 98$\\\hline
   Steel Fatigue 		&  437 &	22& 	$24 \pm 2$ 	& $28 \pm 2$   & $86 \pm 10 $ 	& $27 \pm 2$ & $219$ \\\hline
    \end{tabular}
\end{table}

\begin{figure}[t]
\begin{center}
\includegraphics [width=120mm]{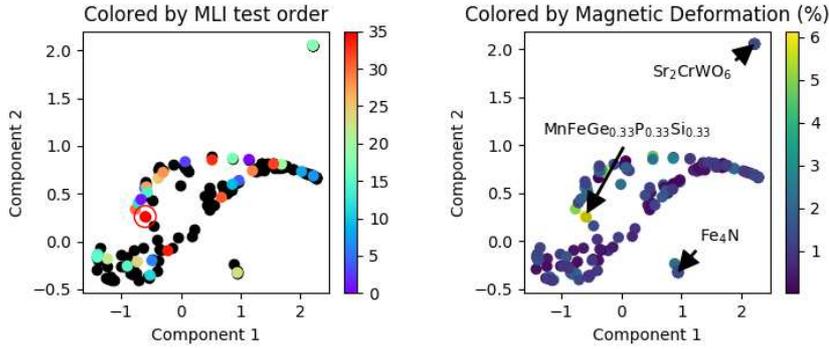}
\end{center}
\caption{
  t-SNE projections of the magnetocalorics data set.
  The axes represent the two components of the t-SNE projection.
  The left plot colors correspond to the order in which the candidates were sampled during a FUELS MLI run.
  The point circled in red was the optimal candidate.
  The black points represent candidates that were not evaluated before the optimal candidate was found.
  The plot on the right is colored by the value of the magnetic deformation for each candidate.
  Representative points from each of the islands, as well as the optimal point, are labeled with their compositions.
} 
\label{magnetocalorics-tsne-fig} 
\end{figure}

One way to visualize the optimization process is through a dimensionality reduction technique called t-SNE~\cite{Maaten2008}.
t-SNE can be used to project the 54-dimensional input vector into 2 dimensions, preserving distances between nearby points as much as possible.
While global distances in t-SNE are not preserved, points that are near to each other in the full-dimensional data set should also be nearby in the t-SNE projection.  
The two dimensions of the t-SNE projection have no physical meaning; their purpose is just to reflect the distance between test candidates in feature space.
t-SNE is analogous to Principal Component Analysis (PCA) in that they both reduce dimensionality, however t-SNE has been shown to be more effective in preserving local distances~\cite{Maaten2008}.
Figure~\ref{magnetocalorics-tsne-fig} shows the t-SNE projection of the magnetocalorics data set and indicates the order in which candidates were evaluated by FUELS MLI, the best-performing SL strategy for this test case.
As this plot shows, the FUELS MLI algorithm explored candidates in all regions of the t-SNE plot before sampling more densely near the optimal point.
Points on both of the ``islands'' were sampled relatively early on.
This behavior is consistent with the tendency of FUELS MLI to explore points with high model prediction uncertainty. 

\subsection{Superconductor Test Case}
\subsubsection{Problem Description}

There is significant interest in developing superconductors with higher critical temperatures.
For this test case, the data set consisted of 546 material candidates whose critical temperatures have been compiled into a publicly accessible Citrination database~\cite{citrination}.
The highest critical temperature of these materials was for Hg-1223 (HgBa$_{2}$Ca$_{2}$Cu$_{3}$O$_{8}$) at 134 K.
The goal of the SL process was to find this optimal candidate using the fewest number of measurements possible.
The inputs were the same 54 real-valued features derived from the chemical formula as were used in the magnetocalorics test case.

\subsubsection{Results}

\begin{figure}[t]
\begin{center}
\includegraphics [width=100mm]{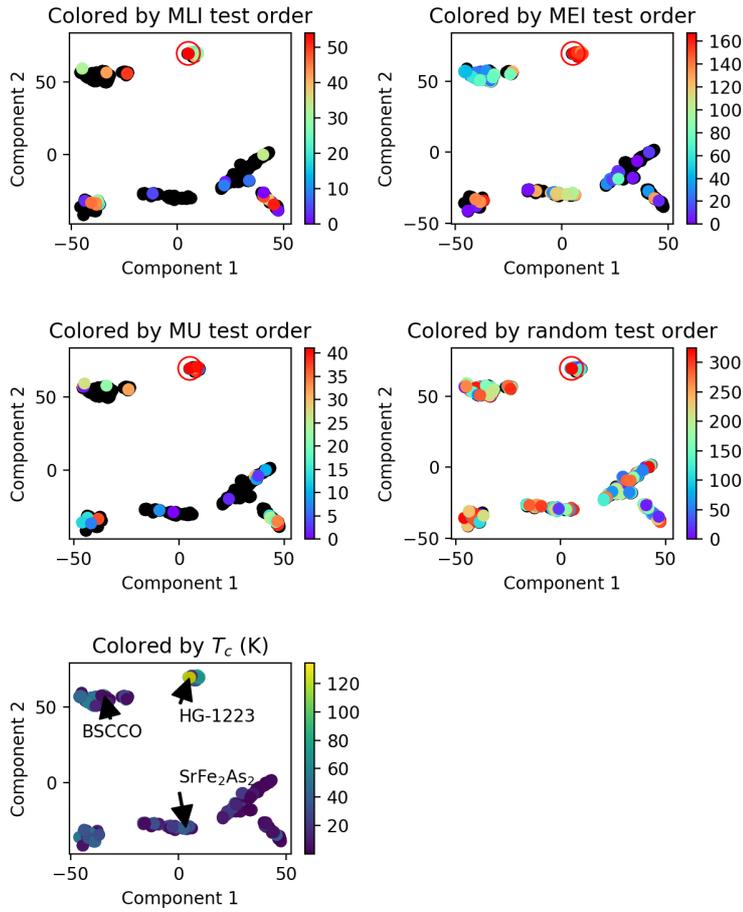}
\end{center}
\caption{
  t-SNE projections of the superconductor data set.
  The axes represent the two components of the t-SNE projection.
  The coloring for the plots corresponds to the FUELS MLI test order, the MEI test order, the MU test order, a random test order, and the value of the critical temperature for each candidate.  
  Points from a couple of the clusters, as well as the optimal point, are labeled with their compositions.
} 
\label{superconductor-tsne-fig} 
\end{figure}

The superconductor data set was substantially larger than the magnetocalorics data set, and it therefore on average required more steps to determine the optimal candidate.
Fig~\ref{steps_fig} shows that FUELS MU required the fewest evaluations in this test case. 
FUELS MU required approximately one fifth and FUELS MLI required approximately one quarter as many evaluations as random guessing.
This performance demonstrates the significant utility and time-savings enabled by SL.
MEI required a slightly larger number of evaluations, perhaps because this strategy does not permit as much exploration.
~\citet{Dehghannasiri2017} also reported that a pure exploitation strategy gave poorer performance in their test case than an exploration-based strategy.
Figure~\ref{superconductor-tsne-fig} shows the t-SNE projection of the FUELS MLI, MEI, MU, and random guessing strategies.
As this plot shows, the random guessing strategy requires the evaluation of a large number of candidates before the optimum is found.
The MEI strategy leads to many nearby candidates being evaluated successively, as indicated by many nearby points having similar colors.  
The coloring for the MLI and MU strategies shows more jumping around, with nearby points often having very different colors.
In cases like this, where the optimal candidate is in a small, isolated cluster, MEI is likely to be less efficient than a more explorative strategy.

\subsection{Thermoelectrics Test Case}
\subsubsection{Problem Description}
In this test case, the data set consisted of 195 materials for which the thermoelectric figures of merit, ZT, as measured at $300 K$, have been compiled into an online Citrination database~\cite{Sparks2016}.  
The inputs to the machine learning algorithm included not only the 54 features calculated from the material formula, but also the semiconductor type (p or n) as well as the crystallinity (\textit{e.g.} polycrystalline or single crystal) of the material. 
 The goal of the optimization was to find the candidate with the highest value of ZT using the fewest number of evaluations.

\subsubsection{Results}

\begin{figure}[t]
\begin{center}
\includegraphics [width=120mm]{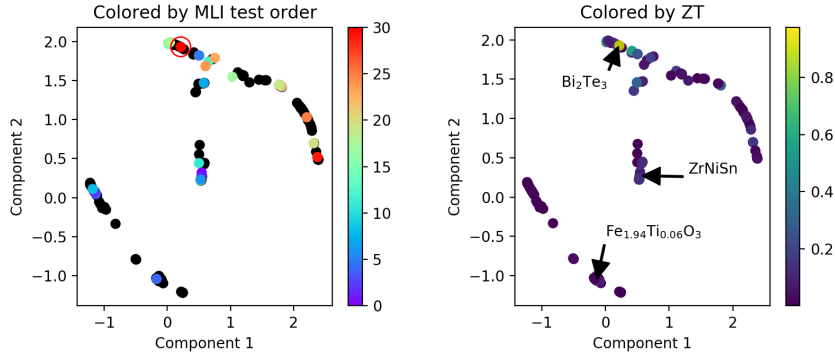}
\end{center}
\caption{
  t-SNE projections of the thermoelectrics data set.
  The axes represent the two components of the t-SNE projection.
  In the left plot, the coloring corresponds to the FUELS MLI test order.  
  In the right plot, the coloring corresponds to the value of ZT at $300 K$ for each candidate.
} 
\label{thermoelectrics-tsne-fig} 
\end{figure}

As is shown in Table~\ref{steps_table}, the three FUELS strategies and COMBO all out-perform random guessing by a significant margin in this test case.  
In particular, the FUELS MU and MLI strategies reduce the mean number of evaluations required by a factor of more than three as compared to random guessing.  
Figure~\ref{thermoelectrics-tsne-fig} shows the t-SNE projection for this test case.  
Perhaps the good performance of all the SL strategies in this test case is due to the fact that the good candidates are clustered near each other in feature space, as indicated by the candidates with high ZT that lie near the optimal candidate in the t-SNE plot. 
 In such cases where the best candidate is near to other good candidates, SL strategies, including greedy strategies, are likely to be significantly more efficient than random guessing.

\subsection{Steel Fatigue Strength Test Case}
\subsubsection{Problem Description}

This test case combined both material composition and process optimization.  
The goal was to find the composition and processing that led to the highest fatigue strength in steel. 
 The data set was based on that of \citet{Agrawal2014}, which included 437 different combinations of steel composition and processing.  
 The features included the fractional composition of 9 different elements (C, Si, Mn, P, S, Ni, Cr, Cu, Mo) as well as 13 processing steps (including tempering temperature, carburization time, and normalization temperature).  
 \citet{Agrawal2014} showed that given these inputs, it was possible to fit a data-driven model that could accurately predict the steel fatigue strength when evaluated via cross-validation.  
 The goal of this test case was to find the combination of the 22 input parameters that led to the candidate with the highest fatigue strength.

\subsubsection{Results}

Figure~\ref{steps_fig} shows that COMBO, FUELS MLI, and FUELS MEI all had very good performance on this test case, finding the optimal set of process and composition parameters in less than 15\% of the number of evaluations as random guessing.  
Interestingly, FUELS MU did not perform well in this case.  
Since FUELS MU is driven by testing those candidates with high uncertainty, it performs well in cases where the optimal candidate is significantly different in some respect from the rest of the candidates.  
MLI and MEI, on the other hand, will fare better when the random forest is able to build an accurate model for the target quantity with the limited data from previously measured candidates.  
Since \citet{Agrawal2014} have already shown that it is possible to use these input features to build an accurate model for the steel fatigue strength, these greedy strategies were able to find the optimal candidate very efficiently in this test case.

\begin{figure}[t]
\begin{center}
\includegraphics [width=120mm]{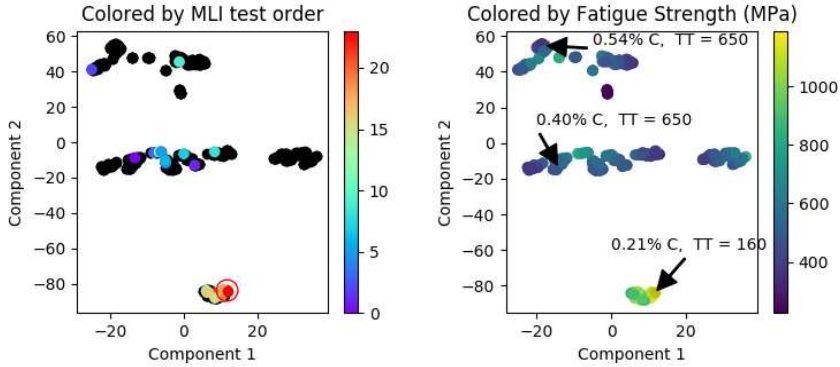}
\end{center}
\caption{
  t-SNE projections of the steel fatigue strength data set.
  The axes represent the two components of the t-SNE projection.
  In the left plot, the coloring corresponds to the FUELS MLI test order.  
  In the right plot, the coloring corresponds to the steel fatigue strength for each candidate.
  Points from a couple of the clusters, as well as the optimal point, are labeled with their percentage carbon content and tempering temperature in Celsius.
} 
\label{steel-tsne-fig} 
\end{figure}

The t-SNE projection for this test case with the MLI strategy is shown in Fig.~\ref{steel-tsne-fig}.  
As this figure shows, most of the candidates with the highest fatigue strengths are grouped together in a cluster characterized by a lower tempering temperature (TT) that in this data set was associated with carburization processing. 
 MLI is able to quickly home in on this cluster and evaluates several candidates in the cluster before finding the optimum.  

\section{Conclusion}\label{ConclusionsSec}

A sequential learning methodology based on random forests with uncertainty estimates has been proposed. 
 The uncertainty was calculated using bias-corrected infinitesimal jackknife and jackknife-after-bootstrap estimates, and was shown to be well-calibrated.  
 This result is significant unto itself, since well-calibrated uncertainty estimates are critical for data-driven models in materials science and other engineering applications.  
 These results represent some of the first evaluations of random forest uncertainty bounds for scientific applications. 
 An implementation of random forests with these uncertainty bounds has been made available through the open source Lolo package~\cite{Lolo}.
 
 The FUELS process has applicability to a wide range of engineering applications with large numbers of free parameters.  
   In this paper, we explored its effectiveness on four test cases from materials science: maximizing the magnetic deformation of magnetocaloric materials, maximizing the critical temperature of superconductors, maximizing the ZT of thermoelectrics, and maximizing fatigue strength in steel. 
   In all of these test cases, the experimental designs were parameterized using between twenty and sixty different features, leveraging the good scaling of FUELS to high-dimensional spaces.
    In all four cases, FUELS significantly out-performed random guessing.  
    While random guessing might seem like a naive benchmark, it should be noted that the data sets in these initial test cases all comprise materials candidates that were thought promising enough to measure.
     Future work will evaluate the impact of SL on a real application for which the optimal candidate is not known \textit{a priori}.

   t-SNE projection was used to enable visualization of the FUELS candidate selections.  
   Three different FUELS strategies were compared: MLI, MEI, and MU.  
   In these initial test cases, MLI consistently had the highest performance.
   MEI struggled in cases where more exploration of the parameter space was important, and MU performed poorly when the random forest model could make accurate predictions after being fit to only a few training points.
   The FUELS approach also compared favorably to the Bayesian optimization COMBO approach, matching its performance in finding the optimal candidate on all four test cases.
   While the COMBO algorithm was designed for scalability to large data sets, it was less computationally efficient than FUELS for these relatively small, high-dimensional data sets.
   While rigorous comparisons of computational efficiency were beyond the scope of this study, in our runs on the steel fatigue strength test case, FUELS was an order of magnitude faster than COMBO per iteration on average in determining the next candidate to test.
   Because the Citrination platform provides publicly-accessible, cloud-hosted machine learning capabilities, the computational efficiency of the experimental design process is important.

   The consistent success of the FUELS strategies in out-performing random guessing underlines the importance and potential impact of optimal experimental design in materials optimization.  
   With experimental efforts representing a bottleneck in the optimization process, it is critical that they be performed in the most efficient manner possible.  
   It is worth noting that the FUELS methodology is equally applicable to both material composition optimization and process optimization.  
   SL provides a framework for minimizing the number of experiments required to identify high-performance materials and optimal processes. 
    It is not our suggestion that SL be used to replace scientific or engineering domain knowledge.  
    Rather, the SL suggestions can be used in supplement to this domain knowledge to provide a quantitative framework to leverage data as it is collected to inform future experiments.

\begin{acknowledgement}
The authors would like to thank S. Wager and T. Covert for their discussions regarding random forest uncertainty estimates.
The authors would also like to thank the rest of the Citrine Informatics team. S. Paradiso and M. Hutchinson acknowledge support from Argonne National Laboratories through contract 6F-31341, associated with the R2R Manufacturing Consortium funded by the Department of Energy Advanced Manufacturing Office.
\end{acknowledgement}
\bibliographystyle{spbasic}    
\bibliography{biblio}
\end{document}